\pdfoutput=1

\documentclass[11pt]{article}

\usepackage{emnlp2021}

\usepackage{times}
\usepackage{latexsym}
\usepackage[T1]{fontenc}
\usepackage[utf8]{inputenc}
\usepackage{microtype}
\usepackage{booktabs, multirow} 
\usepackage{soul}
\usepackage{adjustbox}
\usepackage{changepage,threeparttable} 
\usepackage{tabularx}

\usepackage{amsmath, amsfonts, amssymb,amsthm}
\usepackage{nicefrac}

\usepackage{xcolor}
\definecolor{light-gray}{gray}{0.95}

\usepackage[capitalize]{cleveref}
\usepackage{float}
 \graphicspath{{./figs/}}

\newcommand{\modelsize}[1]{{\tiny\textsubscript{#1}}}

\newcommand{\blank}[0]{\rule{1cm}{0.15mm} }

\newcommand{\mcb}[2]{\multicolumn{#1}{c}{\textbf{#2}}}
\newcommand{\mcbo}[1]{\mcb{1}{#1}}

\newcommand{\mrvert}[2]{\parbox[t]{2mm}{\multirow{#1}{*}{\rotatebox[origin=c]{90}{#2}}}}

\newcommand{\flatpm}[2]{$#1 \pm #2$}
\newcommand{\boldpm}[2]{$\textbf{#1} \pm \textbf{#2}$}

\title{The World of an Octopus: How Reporting Bias Influences a Language Model's Perception of Color}

\author{Cory Paik, Stéphane Aroca-Ouellette,\thanks{*Email has no accent, but includes the hyphen.}  ~Alessandro Roncone \and Katharina Kann \\
  University of Colorado Boulder\\
  \texttt{firstname.lastname@colorado.edu}}
  
\begin{document}
\maketitle
\begin{abstract}
  Recent work has raised concerns about the inherent limitations of text-only pretraining. In this paper, we first demonstrate that \emph{reporting bias}, the tendency of people to not state the obvious, is one of the causes of this limitation, and then investigate to what extent multimodal training can mitigate this issue. To accomplish this, we 1) generate the Color Dataset (CoDa), a dataset of human-perceived color distributions for 521 common objects; 2) use CoDa to analyze and compare the color distribution found in text, the distribution captured by language models, and a human's perception of color; and 3) investigate the performance differences between text-only and multimodal models on CoDa. Our results show that the distribution of colors that a language model recovers correlates more strongly with the inaccurate distribution found in text than with the ground-truth, supporting the claim that reporting bias negatively impacts and inherently limits text-only training. We then demonstrate that multimodal models can leverage their visual training to mitigate these effects, providing a promising avenue for future research.
  
\end{abstract}
\section{Introduction}
Given sufficient scale, language models (LMs)\footnote{In this paper, we use LM to refer to both causal LMs as well as masked LMs.} are able to function as knowledge bases, yielding factoids and relational knowledge across a wide range of topics \cite{petroni2019language,bouraoui-2020-inducing}. However, subsequent work \cite{climbing, bisk2020experience,prost} has raised concerns about the inherent limitations of text-only pretraining. Motivated by these concerns and limitations, we identify and investigate how reporting bias, a concrete and measurable signal, correlates with these limitations and how multimodal training can mitigate these issues.

Grice’s conversational maxim of quantity \cite{grice1975logic} asserts that utterances only contain the required amount of information. This leads to explicit reporting of self-evident knowledge being rare, while less common facts, properties, or events are being reported at disproportionately high frequencies. For example, while most people agree that bananas are typically yellow, the bi-gram ``green banana'' is 332\% more frequent in the Google Books Ngram Corpus \cite{googlengrams} than ``yellow banana''.\footnote{We calculate this number using version 3 from February 2020.} This reporting bias inevitably propagates from corpora to the models trained on them \cite{shwartz2020neural} and affects a variety of concepts. 
One such concept that we would expect to be harmful in downstream applications, is easy to measure, and is solvable via visual input is color. For these reasons, we investigate the relationship between reporting bias and modern LMs' perception of color.

\begin{figure}[t!]
  \centering
\includegraphics[width=\columnwidth]{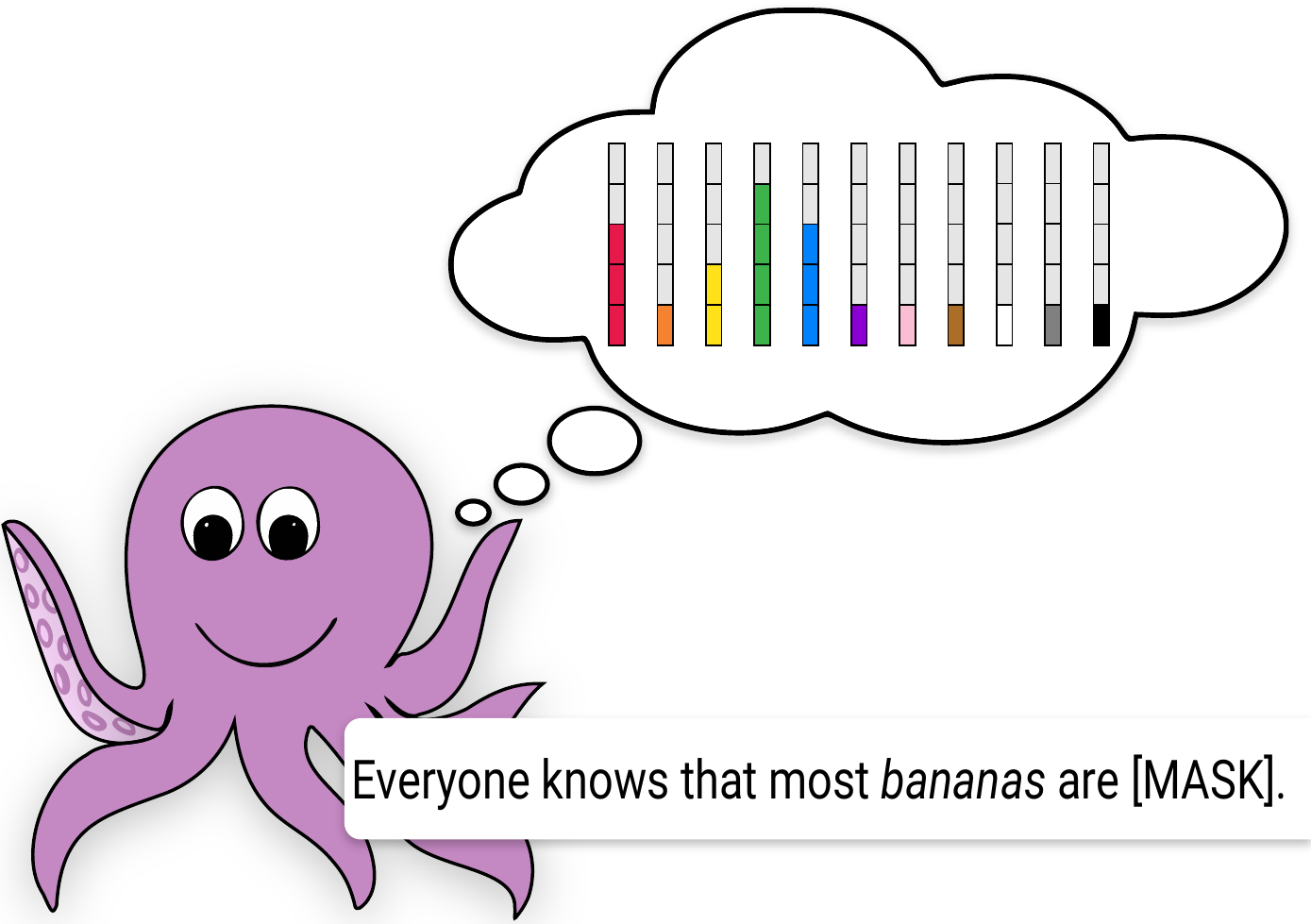}
  \caption{An example prompt from CoDa.}
  \label{fig:example1}
\end{figure}

People's understanding of color is primarily derived from their experience in the world. Every time we interact with an object, we update our understanding of the possible colors that object can take on. Further, we can often apply meaning to the differences: a green banana is unripe, a yellow banana is ideal, and a brown banana may be past its prime. Text-only LMs do not share this embodied experience. Similar to an octopus\footnote{The octopus is a species which has no color photo-receptors and is the protagonist of the thought experiment in \citet{climbing}.} they cannot see colors, and need to rely solely on the inaccurate reporting of colors in text. Thus, we expect the colors LMs associate with objects to differ drastically from a human's perception.

To test this hypothesis, we construct the Color Dataset (CoDa) -- a ground-truth dataset of color distributions for 521 well-known objects via crowdsourcing.
We use this dataset to compare the color distributions found in text and those predicted from LMs, finding that a LM's shortcomings in recovering color distributions correlates with the reporting bias for those objects. Next, we hypothesize that models having access to multiple modalities, specifically vision and text, may be able to partially overcome these shortcoming by grounding the language to their limited visual experiences \cite{bisk2020experience}. To this end, we develop a unified framework for evaluating the color perception of text-only and multimodal architectures. Our results support the hypothesis that multimodal training can mitigate the effect of reporting bias.

\noindent\textbf{Contributions} We make three contributions: 1) We introduce a dataset with human color distributions for 521 well-known objects. 2) We conduct an extensive analysis to identify how reporting bias affects LMs' perception of color. 3) We demonstrate that multimodal training mitigates, but not eliminates, the impact of reporting bias.

\section{CoDa}
\subsection{Dataset Creation}
\begin{table}[t!]\centering \small
\setlength{\tabcolsep}{9.5pt}
\begin{tabular}{lrrr}
\toprule
\mcbo{Dataset} & \mcb{3}{Count (Percentile)} \\
\cmidrule{2-4}
& \mcbo{25\%} & \mcbo{50\%} & \mcbo{75\%} \\
\midrule
Open Images V6 & 2.10K & 3.96K & 11.1K \\
Google Ngrams & 1.63M & 4.64M & 25.4M   \\
Wikipedia     & 2.04K & 10.3K & 38.8K   \\
VQA           & 4     & 25    & 186    \\
\bottomrule
\end{tabular}
\caption{Object frequencies in each domain/dataset after filtering. We report class label statistics for Open Images and $n$-gram frequencies for Google Ngrams, Wikipedia, and VQA prompts. }
\label{tab:freqagg}
\end{table}

\begin{figure*}[t!]
  \centering
\includegraphics[width=.7\linewidth]{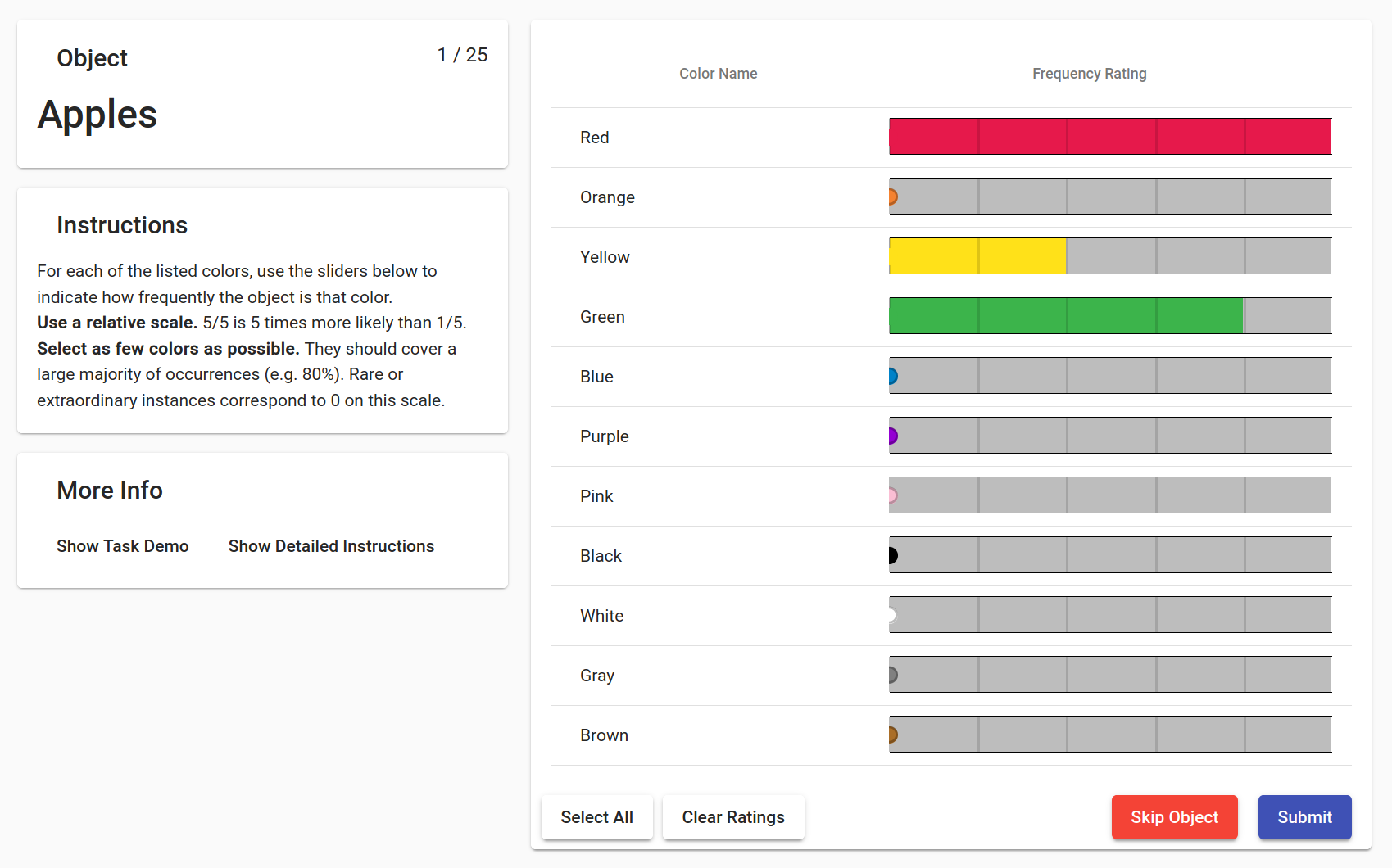}
\caption{Our task UI for data collection on Amazon Mechanical Turk. See \cref{sec:color_ann} for full details.}
\label{fig:mturk1}
\end{figure*}
\paragraph{Object Selection}
To ensure all our models -- and potential future models -- are properly exposed to the objects in our probing dataset, we choose objects which are common in both text and image data. We start with objects from the Open Images dataset \cite{OpenImages} and remove all objects which appear less than 25 times in Wikipedia. For example, we remove “dog bed” as the corresponding bi-gram only appears 19 times. This leaves us with an initial set of 687 objects. 

We then manually filter out all human-related words, such as ``person'' as well as hypernyms such as ``food'', since they are too general to assign specific colors. We also remove transparent objects, such as ``windows'', and objects that are more than two words long, such as ``personal flotation device'' and ``table tennis racket''. This leaves us with our final set of 521 objects. We provide object frequencies from Open Images V6 \cite{OpenImages}, the Google Books Ngram Corpus \cite{googlengrams}, Wikipedia, and VQA \cite{goyal2017vqa} in \cref{tab:freqagg}.

\paragraph{Color Selection}
Following \citet{berlin1969}, we choose the 11 basic color terms of the English language as the colors to be annotated: red, orange, yellow, green, blue, purple, pink, black, white, grey, and brown. 

\paragraph{Color Annotation}
\label{sec:color_ann}
Due to sample bias in image datasets \cite{torralba2011} and the difficulty of matching pixel values to human perception, generating color distributions by counting color frequencies in images is impractical and challenging to verify.\footnote{We attempted an image search paradigm, but challenges such as varied lighting, imperfect segmentation, and the complexity of aligning colors to human perception meant that such a method would still have required human verification.} Thus, in line with our focus on human perception of color as it relates to language (i.e., color terms), we approximate color distributions via human annotation crowdsourced on Amazon Mechanical Turk (MTurk).\footnote{This project went through our institution's ethics review before crowdsourcing was initiated.}

Workers are shown words representing objects and tasked with rating -- on a scale from 1 to 5 -- the frequency with which instances of the objects appear in each of the 11 provided colors.
We set up these tasks as human intelligence tasks (HITs), and provide the workers with instructions, which include an example for how one could label ``grass'' and a concrete list of acceptance and rejection criteria. Each HIT includes 25 objects and is compensated with \$1. 
\cref{fig:mturk1} shows the user interface as presented to an MTurk worker tasked with annotating the object ``apples''. 

Since we choose objects that appear frequently in datasets, we expect people to be familiar with them. However, for the rare cases where an annotator is unsure about an object's color, our interface includes a skip button. The average crowdworker skips 1 object. If an object is not skipped, the average worker completes one annotation in 14 seconds on average. Each object's annotation is normalized to obtain a probability distribution over colors.

A potential side-effect of crowdsourcing annotations is that annotators might choose fewer colors to minimize the time spent on the task. In light of this, we design a labeling interface that balances the time required for labeling a given object as one, many, or all colors. For example, we include a ``Select All'' button and use wide click-optimized sliders. With these changes, we find that, on average, users tend to select 6.2 colors per object. For more details and analysis regarding annotator biases, we refer the reader to \cref{app:annotator_biases}.

\paragraph{Quality Control } For quality control purposes, each HIT includes ``spinach'' as a control object at a random position within the group of objects to annotate. 
This control object serves as a way to flag any submissions which do not follow the instructions or are otherwise not suitable for our purposes.\footnote{Annotators are made aware that control objects with known color distributions are included in the HIT.} We require the rating of ``spinach'' to be more than 50\% green in order to accept the HIT. Rejected HITs are not included in the dataset. This filters out the small number of workers who provide random or blatantly incorrect annotations.

We compute the ground truth as an average over all submitted annotations for a given object. We iteratively filter annotations on a per-object basis if a rating has a Kendall correlation of less than 0 with the current ground truth. This removes 10 annotations that appear to be cases of annotator misinterpretation. For example, one annotator labels ``stop sign'' as being equally red, yellow, and green, likely confusing ``stop sign'' with ``traffic light''.

\begin{table}[th!]\centering\footnotesize
\setlength{\tabcolsep}{4.5pt}
\begin{tabular}{lccccl}
\toprule
\mcbo{Group} & \mcb{1}{All} &\mcb{1}{Train} & \mcb{1}{Val} &\mcb{1}{Test} & \mcbo{Examples}\\
\midrule
Single &  198 &    118 &   39 &    41  & Carrot, Spinach \\
Multi  &  208 &    124 &   41 &    43  & Apple, Street light \\
Any    &  115 &     69 &   23 &    23  & Shirt, Car \\
\midrule
Total  &  521 &    311 &  103 &   107 \\
\bottomrule
\end{tabular}
\caption{CoDa splits by object group.
}
\label{tab:datasetstats}
\end{table}

\paragraph{Object Grouping}
We are investigating the relationship between LMs' knowledge of object colors and reporting bias, the tendency of humans to not state the obvious \cite{grice1975logic}.
We hypothesize that reporting bias will be more severe for objects which have a single typical color, as that color will be implicitly assumed by a listener or reader and, accordingly, will be less frequently stated explicitly. In contrast, objects with a distinct set of several possible colors require explicit descriptions to fully capture the visual characteristics of the object. For example, apples are often described as red or green.  

To test whether objects with different color distributions are impacted by reporting bias differently, we divide the dataset into three categories: single-color objects, multi-color objects, and any-color objects.
We categorize objects using $k$-means clustering with the Jensen-Shannon distance of sorted probabilities. This creates clusters which are color-invariant and based only on the properties of the distributions. We find that this method gives consistent clusters, i.e. the clusters are independent of seeding. We then assign group names semi-manually.\footnote{As there are 3 groups, we can simply mark the ``extreme'' clusters as Single and Any.} 
``Lemon'' is an example of a single color object, where 73\% of the distribution is yellow. ``Wine'' is a multi-color object with 90\% of the distribution falling on red, white, pink, and purple (the last 10\% is yellow). All other objects are any-color objects: they have no clear set of typical colors. Examples of any-color objects are t-shirts, cars, or flowers. More examples are shown in \cref{tab:datasetstats}.

\begin{table}[th!]\footnotesize \centering
\setlength{\tabcolsep}{20pt}
\begin{tabular}{ll}
\toprule
\mcbo{Model Type} & \mcbo{Input} \\
\midrule
Decoder & Most apples are $\mathbf{\left<O\right>}$. \\
Encoder & Most apples are $\mathbf{\left<M\right>}$.  \\
CLIP & A photo of an apple. \\
\bottomrule
\end{tabular}
\caption{Example inputs for different evaluated architectures.}
\label{tab:processing}
\end{table}

\subsection{Templates}
\label{sec:templates}
Text-only corpora and visually-grounded datasets rarely occupy the same domain. To accommodate both, we form a set of templates for each domain. The first is tailored to text-only models, and consists of both plural templates such as ``Most bananas are [MASK].'' and singular templates such as ``This banana is [MASK]''. 

Our second template group is tailored to visually-grounded datasets. We use most of the templates provided by \citet{clip}, which the authors used for finetuning on ImageNet, but exclude templates that inherently point to an unnatural object state, such as ``a photo of a dirty banana''.  Examples for templates are provided in \cref{tab:processing}.  

We recognize that any hand-crafted templates are by nature imperfect. As such, we use all configurations for all models and present the best results per-object for each model to give models ample opportunity to succeed.

\subsection{Data Splits} 
Some of our experiments (cf. \cref{sec:reprprobing}) require a small training set. Thus, CoDa contains training, development and test splits, with 311, 103, and 106 objects respectively. There is no object overlap between the different sets.

\begin{table*}[th!]\centering \small
\begin{tabular}{llrcccc}
\toprule
\mcbo{Dataset} &
\mcbo{Group} & 
\mcbo{Freq} & 
\mcbo{Spearman $\rho \uparrow$} & 
\mcbo{Kendall's $\tau \uparrow$} &  
\mcbo{Acc@1 $\uparrow$} & 
\mcbo{$\mathbf{D_{JS}}  \downarrow$} \\
\midrule 
Google Ngrams & Single &   $5.60$ &      \flatpm{41.7}{27.8} &       \flatpm{35.3}{24.5} &  $\textbf{43.9}$ &  \flatpm{0.27}{0.16} \\
    & Multi &   $9.69$ &      \boldpm{47.1}{26.6} &       \boldpm{38.1}{22.2} &           $30.3$ &  \flatpm{0.23}{0.12} \\
    & Any &  $20.26$ &      \flatpm{43.5}{30.7} &       \flatpm{34.3}{25.0} &           $33.9$ &  \boldpm{0.15}{0.10} \\
\midrule 
Wikipedia & Single &   $1.51$ &      \flatpm{26.5}{30.2} &       \flatpm{22.2}{26.3} &  $\textbf{25.3}$ &  \flatpm{0.37}{0.17} \\
    & Multi &   $1.85$ &      \flatpm{29.4}{31.9} &       \boldpm{23.9}{27.0} &           $23.6$ &  \flatpm{0.31}{0.16} \\
    & Any &   $3.00$ &      \boldpm{30.9}{31.5} &       \flatpm{23.8}{25.6} &           $19.1$ &  \boldpm{0.23}{0.15} \\
\midrule 
VQA & Single &   $0.73$ &      \flatpm{27.4}{37.8} &       \flatpm{25.4}{35.3} &           $16.7$ &  \flatpm{0.38}{0.23} \\
    & Multi &   $2.17$ &      \boldpm{35.7}{34.3} &       \boldpm{31.7}{30.9} &           $21.2$ &  \flatpm{0.35}{0.20} \\
    & Any &   $2.64$ &      \flatpm{33.7}{33.6} &       \flatpm{28.1}{28.7} &  $\textbf{27.8}$ &  \boldpm{0.29}{0.17} \\
\bottomrule
\end{tabular}

\caption{Correlation metrics between the $n$-gram frequencies reported in different datasets and the ground truth distributions collected from human annotators. Single, Multi, and Any indicate sets of objects that are frequently a single color, between two to four colors, or could be any color, respectively. We aggregate by object and report the mean $\pm$ standard deviation for each metric across the objects of that group.    
}
\label{tab:rb1}
\end{table*}
\section{Reporting Bias}

\subsection{Background}
As previously stated, Grice’s conversational maxim of quantity manifests as \textit{reporting bias} -- i.e., people not usually stating obvious facts or properties --, and impacts nearly all datasets that contain text.

Reporting bias has been studied in the context of both NLP and image captioning. \citet{gordon2013reporting} perform a quantitative analysis using n-gram frequencies from text, finding this phenomenon particularly relevant to internet text corpora. \citet{shwartz2020neural} extend these experiments to pretrained models such as Bert \citep{bert} and RoBERTa \citep{roberta}. Similar to our work, they analyze color attribution of the form ``The \blank banana is tasty.'' However, 
their ground truth is extracted from Wikipedia bi-grams and, thus,  suffers from reporting bias itself. In contrast, we circumvent this problem by collecting the ground truth in CoDa directly from humans.

\subsection{Reporting Bias in Text}
Our hypothesis is that pretrained LMs inherit reporting bias with respect to colors from their training data. Thus, prior to our main experiments, we investigate if, in fact, reporting bias exists in large general text corpora.  
We analyze the Google Books Ngram Corpus \cite{googlengrams} and Wikipedia. Specifically, we look at all bi-grams and tri-grams containing a color followed by an object in our dataset. 

Let us denote the count of the $n$-gram $x_1 \ldots x_n$ as $\phi(x_1, \ldots, x_n )$. We then define the relative frequency with which each object $o$ appears with a color $c$ as: 
\begin{equation}
\textrm{Freq}(o) = \frac{100}{\phi(o)} \sum_{c \, \in \, C} \phi(c, o) 
\end{equation}
We further define the 
probability of an object being of color $c^*$ as:
\begin{equation}
P(c^* \,  | \, o) = \frac{\phi(c^*, o)}{\sum_{c \, \in \, C} \phi(c, o) }
\end{equation}
The results of these experiments are reported in \cref{tab:rb1}. The frequency column supports our hypothesis that objects with one typical color are  less frequently described as being of any color than those with multiple typical colors or where any color is possible. In all metrics excluding Acc@1, the text-retrieved color distributions are more strongly correlated with the ground truth for multi and any colored objects than for single-colored objects.\footnote{Acc@1 is not directly comparable across object groups, see \cref{sec:metrics} for details.}

\section{Experimental Setup}
\begin{table}[t!]\centering \footnotesize
\begin{tabular}{llc}
\toprule
\mcbo{Model} & \mcbo{Sizes} & \mcbo{Multimodal }\\
\midrule
GPT-2    & B, M, L, XL    \\
RoBERTa  & B, L           \\
ALBERT V1 & B, L, XL, XXL \\
ALBERT V2 & B, L, XL, XXL \\
CLIP & ViT-B/32, RN50,  & \checkmark \\
     & RN50x4, RN101          \\
\bottomrule
\end{tabular}

\caption{Summary of evaluated models. }
\label{tab:models}
\end{table}
\subsection{Zero-shot Probes}
We first probe LMs in a zero-shot fashion using a set of templates (see \cref{sec:templates}). Each template has a [MASK] where the color should appear. For models trained using a causal language modeling objective, we run the models over each template eleven times, each time with a different color replacing the [MASK] token. Following \citet{blimp}, we select the sentence with the highest probability. For models trained using a masked language modeling objective, we filter the output vocabulary to only include the eleven color choices and normalize to obtain a probability distribution.

\subsection{Representation Probes}
\label{sec:reprprobing}
Many current multimodal architectures are optimized for multimodal evaluation and have complex shared embedding spaces, which makes it challenging to compare to text-only models.
However, recent developments such as CLIP \cite{clip} and ALIGN \cite{align} show promising results in connecting images and text via contrastive pretraining on large unlabeled corpora, while still maintaining separate text and image models. We focus on probing multimodal models which follow these architecture decisions. 
Since they have not been trained on a language modeling objective, zero-shot probing is not viable on these models. To overcome this and enable comparison to text-only models, we freeze the base model and use part of our dataset to train a MLP to extract color distributions from the frozen representations.

Given pretrained representations, we would like the performance of a model to consist of 2 parts: final quality (in our case distribution correlations), and the amount of effort to get that quality from the representations. This is possible by formulating the task as \textit{efficiently} learning a model from representations to color distributions. Following \citet{whitney2021evaluating,voitamdl}, we conduct our experiments for representation probing in a loss-data framework using minimum description length (MDL), surplus description length (SDL), and $\varepsilon$ sample complexity ($\varepsilon SC$).   
We split the training set into 10 subsets spaced logarithmically from 1 to 311 objects, and report averages over 5 seeds.

\begin{table*}[th!]\centering\footnotesize
\begin{tabular}{llcccrrr}
\toprule
\mcb{1}{Model} & \mcb{1}{Group} & \mcb{1}{Spearman $\rho \uparrow$} &  \mcb{1}{Kendall's $\tau \uparrow$} & 
\mcb{1}{Acc@1 $\uparrow$} &  \mcb{1}{$\mathbf{D_{JS}} \downarrow$} &
$\Delta \rho \uparrow$  & $\Delta \tau \uparrow$\\
\midrule
GPT-2 & Single &      \flatpm{40.3}{26.6} &       \flatpm{33.6}{22.1} &  $\textbf{40.4}$ &  \flatpm{0.39}{0.07} &                $-0.55$ &                $-1.01$ \\
       & Multi &      \flatpm{44.8}{20.9} &       \flatpm{36.5}{16.8} &           $29.8$ &  \flatpm{0.26}{0.06} &                $-1.49$ &                $-1.05$ \\
       & Any &      \boldpm{48.1}{25.1} &       \boldpm{38.2}{20.2} &           $40.0$ &  \boldpm{0.09}{0.04} &        $\textbf{5.29}$ &        $\textbf{4.46}$ \\
\midrule
RoBERTa & Single &      \flatpm{47.8}{24.7} &       \flatpm{40.1}{20.8} &  $\textbf{42.9}$ &  \flatpm{0.28}{0.11} &                 $7.17$ &                 $5.69$ \\
       & Multi &      \flatpm{50.2}{23.8} &       \flatpm{41.0}{19.5} &           $33.2$ &  \flatpm{0.19}{0.08} &                 $4.57$ &                 $4.01$ \\
       & Any &      \boldpm{52.5}{23.5} &       \boldpm{42.0}{19.5} &           $36.5$ &  \boldpm{0.10}{0.06} &        $\textbf{9.97}$ &        $\textbf{8.26}$ \\
\midrule
ALBERT & Single &      \flatpm{43.7}{24.4} &       \flatpm{36.4}{20.6} &           $34.3$ &  \flatpm{0.30}{0.11} &                 $2.69$ &                 $1.55$ \\
       & Multi &      \flatpm{44.6}{19.1} &       \flatpm{36.1}{15.5} &           $26.9$ &  \flatpm{0.22}{0.07} &                $-1.53$ &                $-1.27$ \\
       & Any &      \boldpm{48.2}{21.4} &       \boldpm{38.2}{17.2} &  $\textbf{35.7}$ &  \boldpm{0.11}{0.05} &        $\textbf{5.07}$ &        $\textbf{4.22}$ \\
\bottomrule
\end{tabular}

\caption{LM results when probed in a zero-shot setting. Single, Multi, and Any indicate sets of objects that are frequently of a single color, between two to four colors, or could be any color, respectively. All correlation coefficients ($\rho, \tau$) are multiplied by 100. For each object, we take the prediction from the template with the highest $\tau$ correlation. We then aggregate by object and report the mean $\pm$ standard deviation over objects of that group. We report the results from the best model from each architecture; for results on a per-model basis, see \cref{tab:lmzeroshot:all}.}
\label{tab:lmzeroshot}
\end{table*}

\subsection{Models}
We probe object-color probabilities in 14 pretrained text-only models as well as four versions of CLIP \cite{clip}. The text-only models are varied configurations of GPT-2 \cite{gpt2}, RoBERTa \cite{roberta}, and ALBERT \cite{albert}; cf. \cref{tab:models} for the full set.  
We use Huggingface's \citep{huggingface} pretrained models for all text-only models and the official implementation of CLIP.\footnote{\href{https://github.com/openai/CLIP}{github.com/openai/CLIP}}

\subsection{Metrics}
\label{sec:metrics}
In order to obtain as comprehensive a picture as possible, we report a variety of metrics when applicable, including: top-1 accuracy, Spearman rank order correlation $\rho$, Kendall rank correlation $\tau$, and Jensen-Shannon divergence $D_{JS}$ for each model and each set of objects. Each of these metrics highlight slightly different aspects of performance on the task. 

Top-1 accuracy (Acc@1) is the frequency with which models can correctly identify the most frequent color of an object. This is useful for comparing models, but not directly interpretable across object groups as it inherently favors objects that can take on few colors.
Spearman's $\rho$ is sensitive to outliers, so it highlights the extreme mistakes, while Kendall's $\tau$ is more robust to such changes.
Jensen-Shannon divergence measures the similarity between 2 distributions. 

Spearman's $\rho$  and Kendall's $\tau$ are within the range of $[-1, 1]$, with -1 being negatively correlated and 1 being perfectly correlated.\footnote{We multiply by 100 in all tables for legibility.}
We additionally define $\Delta \rho$ and $\Delta \tau$ correlation difference measures defined on the interval [-100, 100], to compare model predictions to $n$-gram frequency predictions. This measures the difference in correlation between $n$-gram frequency predictions and a model's probability distribution, where -100 indicates degraded correlation, 0 equals perfect correlation, and 100 indicates improved correlation with the ground truth as compared to the relative n-gram frequencies. In the context of reporting bias, $\Delta \rho$ and $\Delta \tau$ can be interpreted as measures of bias amplification or mitigation for negative and positive values, respectively. 

We additionally define an average of the two correlation metrics as ``Avg. Correlation''. When using this metric, we first compute $\frac{\rho+\tau}{2}$ for a specific object and perform all other aggregations in the same way as for the other metrics.

\begin{figure}[th!] \centering
\includegraphics[width=\columnwidth]{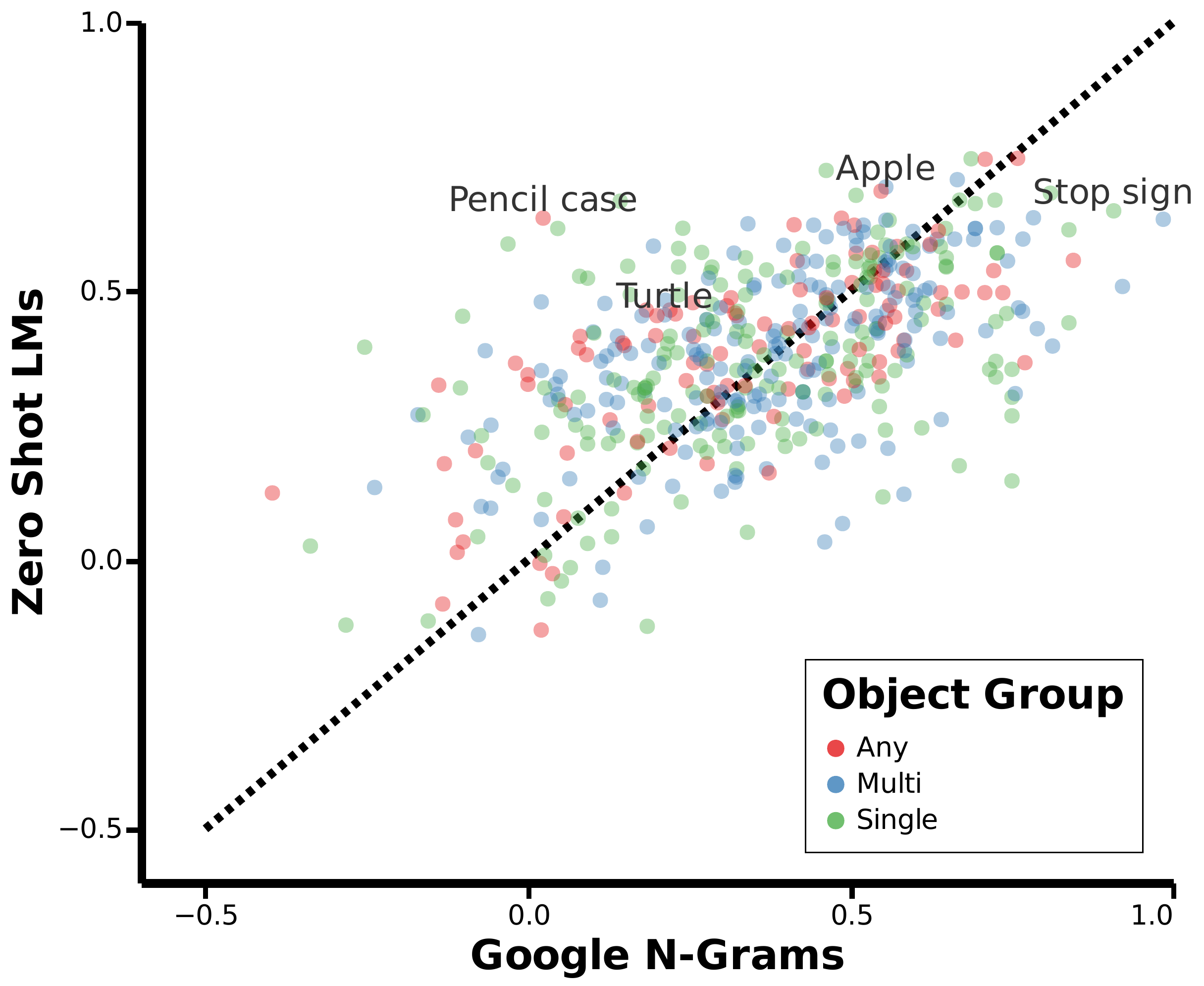}
\caption{Correlation between n-gram frequency and LM performance for single, multi, and any color objects. X and Y axes are Kendall's $\tau$ correlation between n-gram frequency and ground truth and LM predictions with ground truth respectively. Each point corresponds to a single object in our dataset. LM correlation is averaged over the top models for each architecture. The dotted line $y$=$x$ corresponds to to perfect correlation.
}
\label{fig:zsngrams}
\end{figure}

\section{Results}
\subsection{Zero-Shot Probes}
The results of LMs when probed in a zero-shot setting, provided in \cref{tab:lmzeroshot}, clearly demonstrate that LMs perform worse on single-color objects and perform better on objects that can take on a range of colors. Furthermore, correlations are relatively low for all objects and models. This demonstrates that colors are generally challenging for state-of-the-art pretrained LMs.

\begin{figure*}[th!]\centering
\includegraphics[width=\columnwidth]{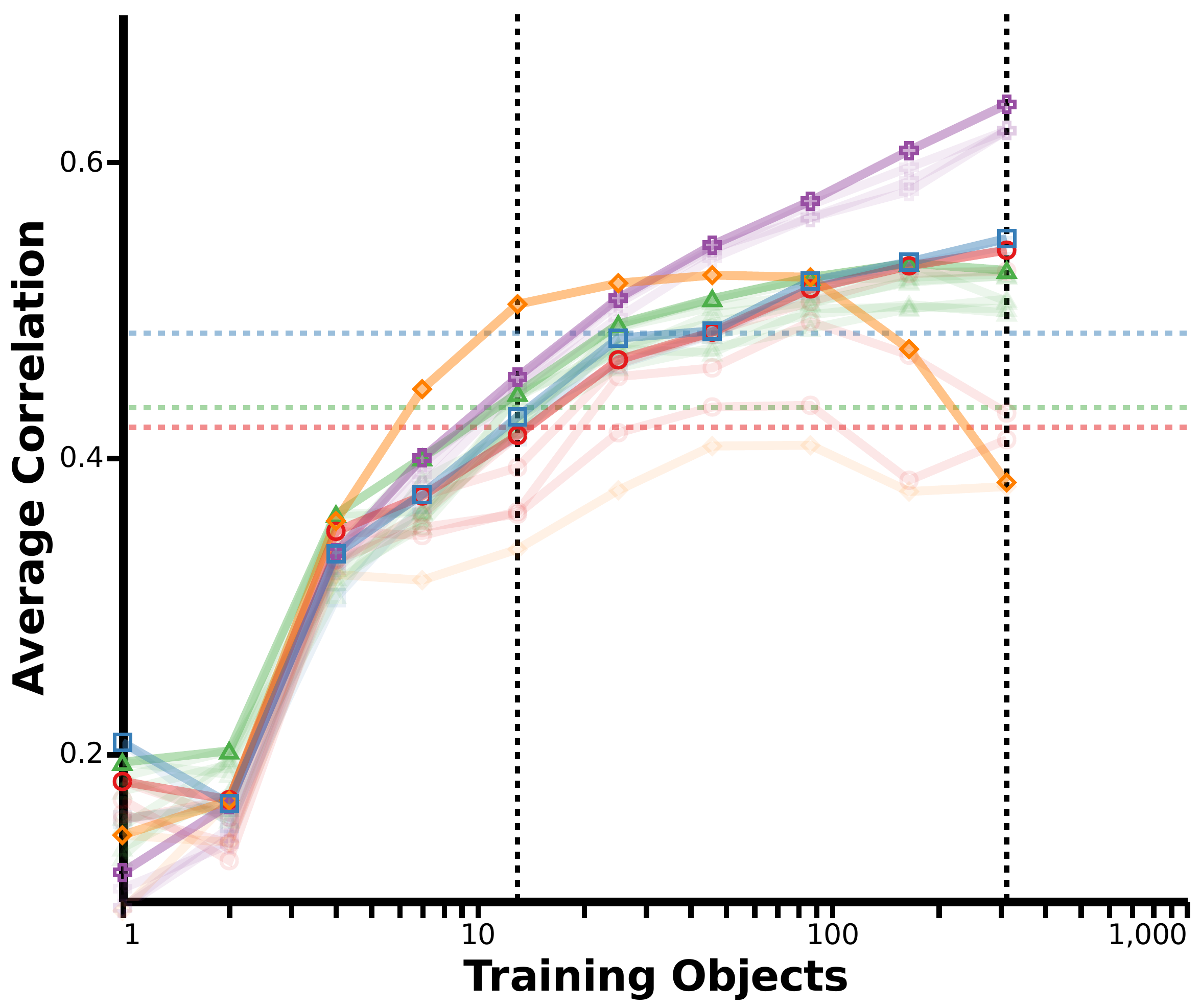}
\includegraphics[width=\columnwidth]{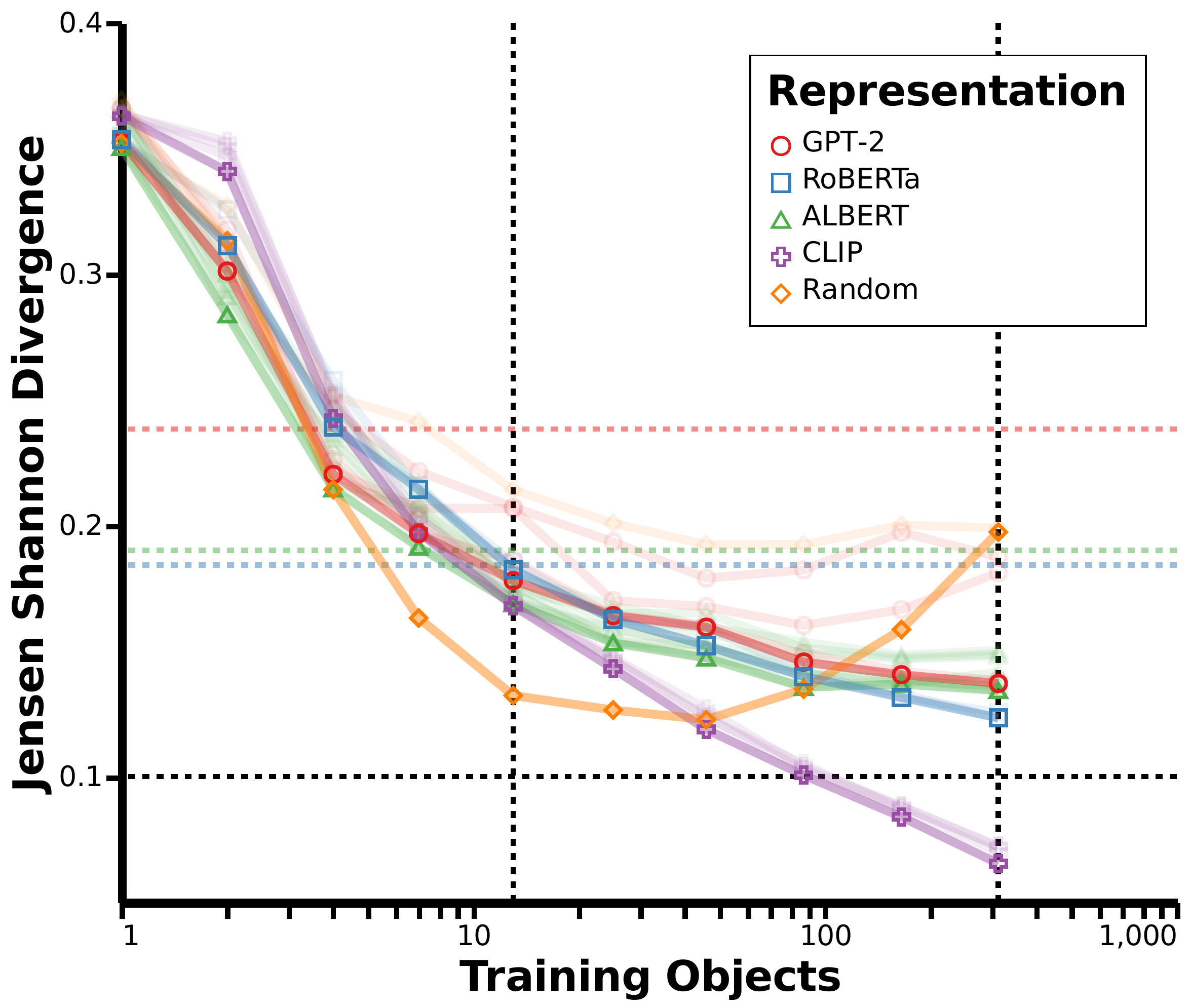}
\caption{Representation probing results for unseen objects with varying amounts of data, averaged over 5 seeds. The main lines are the best model from those of the same type (e.g., RoBERTa\modelsize{BASE} and RoBERTa\modelsize{LARGE}), and the translucent lines are the per-model averages. Dotted lines represent best zero-shot performance for each model. The ``Random'' group consists of a randomly initialized RoBERTa and CLIP. The black dotted lines correspond to $\epsilon$ and $n$ in \cref{tab:reprlossdata}. Left: Average of Spearman's $\rho$ and Kendall's $\tau$. Right: Jensen-Shannon divergence.} 
\label{fig:reprlossdata}
\end{figure*}

\subsection{Reporting Bias and Model Accuracy}
\Cref{fig:zsngrams} compares the correlation between $n$-gram frequency and zero-shot LM performance. The identity line represents a theoretical perfect correlation between how well $n$-gram frequency correlates with our ground truth and LM predictions.\footnote{That is, where LMs directly reflect $n$-gram frequencies.} Any points above the identity line represent cases where LMs seem to \textit{mitigate} reporting bias -- their predictions are closer to ground truth, and points below the line represent cases where LMs \textit{amplify} reporting bias -- their predictions are further from ground truth. When averaged across all models (see \cref{app:zeroshotresults} for the full list of results) zero-shot LMs amplify the reporting bias of single-color objects by 5.23\% on average, and 6.26\% for multi-color objects. 
For any-color objects, we find a slight mitigation of 0.21\% on average.

\cref{tab:rb_lm} aggregates and combines results from \cref{tab:rb1,tab:lmzeroshot} and elucidates two main points on the effect of reporting bias on a LM's perception of color. First, the color distributions of LMs correlate more strongly with reporting bias-affected text than with a human's perception of color. Second, single-colored objects are the most affected by reporting bias, and the objects LMs struggle the most on. These results indicate that, in line with our hypothesis, LMs are negatively impacted by reporting bias. Further, because reporting bias is innate to human communication and due to the enormous amount of text required for modern LMs, it is infeasible to eliminate reporting bias from all training data. This entails -- in support of the arguments in \citet{climbing} and \citet{bisk2020experience} -- that language understanding abilities are naturally limited by text-only training. 

\begin{table}[t!]\centering
\begin{tabular}{lrrr}
\toprule
&& \mcb{2}{Avg. Correlation $\uparrow$}\\
\cmidrule(lr){3-4}
\mcbo{Group} & \mcbo{Freq.}  & \mcbo{Humans} & \mcbo{Ngrams} \\
\midrule
Single &  $5.60$ & \flatpm{40.1}{22.3} &  \flatpm{63.0}{18.1} \\
Multi  &  $9.69$ & \flatpm{42.2}{20.5} &  \flatpm{63.1}{17.5} \\
Any    & $20.26$ & \flatpm{42.9}{22.5} &  \boldpm{63.4}{16.2} \\
\bottomrule
\end{tabular}

\caption{\textbf{LM predictions have higher correlation with n-gram frequencies}.
Here we compare the average correlation between LM predictions and two sources of ``ground truth''; one collected from human annotators and one computed from n-gram frequencies. 
Single, Multi, and Any indicate sets of objects that are frequently of a single color, between two to four colors, or could be any color, respectively. The ``Freq.'' column indicates the frequency n-grams containing these objects also have one of the eleven colors. }
\label{tab:rb_lm}
\end{table}

\begin{table}[t!]\centering \footnotesize
\begin{tabular}{llrrrr}
\toprule
    &  & \mcbo{GPT-2}    & \mcbo{RoBERTa}   & \mcbo{CLIP} \\
n & {} & \scriptsize{L} & \scriptsize{B}  & \scriptsize{ViT-B/32}  \\
\midrule
13  & $\mathbf{D_{JS}}$                  &    0.178 &     0.185 &   0.168 \\
    & MDL                                &    2.80  &     2.95  &       2.79 \\
    & SDL, $\varepsilon$=0.1             &  > 1.50  &   > 1.65  &     > 1.49 \\
    & $\varepsilon$SC, $\varepsilon$=0.1 &    > 13  &      > 13 &       > 13 \\
    & Avg Corr.                          &    40.7  &      42.7 &       45.5 \\
311 & $\mathbf{D_{JS}}$                  &   0.137  &     0.123 &  \textbf{0.065} \\
    & MDL                                &   45.07  &     42.08 &  \textbf{27.22} \\
    & SDL, $\varepsilon$=0.1             & > 13.97  &   > 10.98 &  \textbf{2.43} \\
    & $\varepsilon$SC, $\varepsilon$=0.1 &    > 311 &     > 311 &  \textbf{165}   \\
    & Avg Corr.                          &    54.0  &      54.9 &  \textbf{63.9}  \\
\bottomrule
\end{tabular}

\caption{Estimated measures of representation quality for the best model of each architecture. }
\label{tab:reprlossdata}
\end{table}
\subsection{Representation Probes}
\cref{fig:reprlossdata} shows the average correlation and Jensen-Shannon divergence for unseen objects as a function of the number of training objects. Note that with 14 objects, all models surpass zero-shot performance in terms of Jensen-Shannon divergence. With enough training objects, we observe similar ranking patterns observed in the zero-shot setting for text-only models. However, the advantage of this approach is that we can additionally include multimodal architectures. 

The results from these experiments demonstrate that multimodal models outperform text-only models at recovering color distributions. They manage to do so even though the performance of multimodal models is often lower on classic NLP tasks \cite{vokenization} and many multimodal datasets are even more prone to reporting bias in text \cite{mirsha2016seeing,emiel2018stereotyping,burns2018women}. This further support the arguments in \citet{bisk2020experience} that understanding concepts requires experiencing them in their natural form. 

\section{Limitations}
While our work identifies issues with text-only training and  motivates the use of multimodal signals during pretraining, in this section we outline some limitations of our approach.

First, a number of recent papers have highlighted potential limitations of probing LMs in certain ways \cite{zhang-bowman-2018-language,whitney2021evaluating}. 
While we acknowledge that probing does not provide a full picture of the capabilities of LMs, our hypothesis was supported by a range of different results from different approaches. In future work, we hope to leverage research \citep{bouraoui-2020-inducing,jiang-etal-2020-know} that demonstrates effective methods for automatically producing templates optimized for specific models. In the current state, we cannot and do not state exactly what LMs do and do not capture, rather we use our results to uphold and strengthen our original hypothesis that reporting bias hinders performance and that multimodal signals can help mitigate this problem.

Second, the bi-gram/tri-gram approach we use to quantify reporting bias only approximates the full set of object-color instances. To be more exact, a dependency parser would have to be run on every dataset. 

Finally, although our results motivate the use of multimodal signals during pretraining, there are still challenges to overcome.
As discussed by \citet{vokenization}, the performance of multimodal models on classic NLP tasks often does not reflect the inherent advantages of these architectures, and many multimodal dataset are even more prone to reporting bias in text \cite{mirsha2016seeing,emiel2018stereotyping,burns2018women}. 
Further, while a visual signal is able to better impart a sense of color, it is not enough to endow models with the meaning behind those colors. Humans easily learn that a green banana is not yet ripe, and that a brown banana is past its prime. For models to obtain this level of knowledge and reasoning they will likely require training signals from more modalities, and potentially fully embodied experiences.

\section{Related Work}

\noindent\textbf{Color-Object Relationships} Preexisting word association datasets often include object-color relationships as either having multiple equally likely pairings \cite{BATS,browncorpus}, or as probabilistic cue-target pairs \citep{nelson-2004-university}. Others such as \citet{CLSB} take a norm completion approach, wherein participants are tasked with generating attributes given some concept. One can then extract the object-color relationships by counting the number of participants who reported a given color.

However, the resulting ``distribution'' is an aggregate count over individuals, and does not necessarily reflect the distribution from the eyes of a single observer. Thus, previous research into LMs as knowledge bases has not been able to fully explore the extent to which they know color \cite{rodriguez2020word,shwartz2020neural}.

Previous work has shown the importance of color in visual perception and object recognition \cite{rosenthal-2018-color,gegenfurtner-2000-sensory}. More recently \citet{teichmann2020influence} use time resolved neural imaging data to demonstrate how the typicality of object-color relationships influences object representations in visual processing.

\paragraph{Probing LMs} 
A wide range of papers have probed LMs in a zero-shot fashion by looking at how they fill in a [MASK] token in handcrafted \cite{weir2020probing,petroni2019language,jiang-etal-2020-know,BERTisNot,NumerSense} or automatically generated \cite{bouraoui-2020-inducing,jiang-etal-2020-know} template sentences. Others, such as \citet{blimp} compare perplexities between minimal pairs of sentences.
A different approach is to analyze the representation quality of LMs for linguistic tasks by training a simple MLP on pretrained model representations \citep{da2019cracking,lin2019open}. However, \citet{zhang-bowman-2018-language} demonstrate that the procedure of training an additional classifier may distort the results.
An alternative approach introduced by \citet{voitamdl} is information-theoretic probing with MDL. This method builds on standard probing classifiers by not only measuring the final performance, but additionally measuring the amount of effort required to achieve that performance.

\paragraph{Probing Multimodal LMs} Often multimodal LMs are used in the domain of visual question answering, where, given an image, the model is asked a question about concepts in the image \cite{goyal2017vqa,hudson2018gqa}. While it is often possible to simply use the text-only portion of these models for other tasks, this often leads to poor performance on solely language-based tasks \citep{vokenization}.

\section{Conclusion}
In this paper we investigate how reporting bias negatively effects a LM's perception of color. We do so by first creating CoDa, a dataset of 521 human-perceived color distributions for common objects. We then utilize this dataset to demonstrate that text-only models are inherently limited because of reporting bias. Subsequently, we show that multimodal training mitigates these issues. Overall, our results support the claims in \citet{climbing} and \citet{bisk2020experience} that text-only training is insufficient for language understanding and motivate further research on how to best employ multimodal training signals.

\section*{Acknowledgments}
We would like to thank the members of CU Boulder's NALA Group for their feedback on this work. We would also like to thank the reviewers for taking the time to provide insightful questions and feedback. 

\bibliography{main}

\begin{thebibliography}{45}
\expandafter\ifx\csname natexlab\endcsname\relax\def\natexlab#1{#1}\fi

\bibitem[{A.~Rodriguez and Merlo(2020)}]{rodriguez2020word}
Maria A.~Rodriguez and Paola Merlo. 2020.
\newblock \href {https://doi.org/10.18653/v1/2020.conll-1.30} {Word
  associations and the distance properties of context-aware word embeddings}.
\newblock In \emph{Proceedings of the 24th Conference on Computational Natural
  Language Learning}, pages 376--385, Online. Association for Computational
  Linguistics.

\bibitem[{Aroca-Ouellette et~al.(2021)Aroca-Ouellette, Paik, Roncone, and
  Kann}]{prost}
St{\'e}phane Aroca-Ouellette, Cory Paik, Alessandro Roncone, and Katharina
  Kann. 2021.
\newblock \href {https://doi.org/10.18653/v1/2021.findings-acl.404} {{PROST}:
  {P}hysical reasoning about objects through space and time}.
\newblock In \emph{Findings of the Association for Computational Linguistics:
  ACL-IJCNLP 2021}, pages 4597--4608, Online. Association for Computational
  Linguistics.

\bibitem[{Bender and Koller(2020)}]{climbing}
Emily~M. Bender and Alexander Koller. 2020.
\newblock \href {https://doi.org/10.18653/v1/2020.acl-main.463} {Climbing
  towards {NLU}: {On} meaning, form, and understanding in the age of data}.
\newblock In \emph{Proceedings of the 58th Annual Meeting of the Association
  for Computational Linguistics}, pages 5185--5198, Online. Association for
  Computational Linguistics.

\bibitem[{Berlin and Kay(1969)}]{berlin1969}
Brent Berlin and Paul Kay. 1969.
\newblock \emph{Basic Color Terms: Their Universality and Evolution}.
\newblock University of California Press.

\bibitem[{Bisk et~al.(2020)Bisk, Holtzman, Thomason, Andreas, Bengio, Chai,
  Lapata, Lazaridou, May, Nisnevich, Pinto, and Turian}]{bisk2020experience}
Yonatan Bisk, Ari Holtzman, Jesse Thomason, Jacob Andreas, Yoshua Bengio, Joyce
  Chai, Mirella Lapata, Angeliki Lazaridou, Jonathan May, Aleksandr Nisnevich,
  Nicolas Pinto, and Joseph Turian. 2020.
\newblock \href {https://doi.org/10.18653/v1/2020.emnlp-main.703} {Experience
  grounds language}.
\newblock In \emph{Proceedings of the 2020 Conference on Empirical Methods in
  Natural Language Processing (EMNLP)}, pages 8718--8735, Online. Association
  for Computational Linguistics.

\bibitem[{Bouraoui et~al.(2020)Bouraoui, Camacho{-}Collados, and
  Schockaert}]{bouraoui-2020-inducing}
Zied Bouraoui, Jos{\'{e}} Camacho{-}Collados, and Steven Schockaert. 2020.
\newblock \href {https://aaai.org/ojs/index.php/AAAI/article/view/6242}
  {Inducing relational knowledge from {BERT}}.
\newblock In \emph{The Thirty-Fourth {AAAI} Conference on Artificial
  Intelligence, {AAAI} 2020, The Thirty-Second Innovative Applications of
  Artificial Intelligence Conference, {IAAI} 2020, The Tenth {AAAI} Symposium
  on Educational Advances in Artificial Intelligence, {EAAI} 2020, New York,
  NY, USA, February 7-12, 2020}, pages 7456--7463. {AAAI} Press.

\bibitem[{Burns et~al.(2018)Burns, Hendricks, Darrell, and
  Rohrbach}]{burns2018women}
Kaylee Burns, Lisa~Anne Hendricks, Trevor Darrell, and Anna Rohrbach. 2018.
\newblock \href {https://arxiv.org/abs/1803.09797} {Women also snowboard:
  Overcoming bias in captioning models}.
\newblock \emph{ArXiv preprint}, abs/1803.09797.

\bibitem[{Da and Kasai(2019)}]{da2019cracking}
Jeff Da and Jungo Kasai. 2019.
\newblock \href {https://doi.org/10.18653/v1/D19-6001} {Cracking the contextual
  commonsense code: Understanding commonsense reasoning aptitude of deep
  contextual representations}.
\newblock In \emph{Proceedings of the First Workshop on Commonsense Inference
  in Natural Language Processing}, pages 1--12, Hong Kong, China. Association
  for Computational Linguistics.

\bibitem[{Devereux et~al.(2013)Devereux, Tyler, Geertzen, and Randall}]{CLSB}
Barry Devereux, Lorraine Tyler, Jeroen Geertzen, and Billi Randall. 2013.
\newblock \href {https://doi.org/10.3758/s13428-013-0420-4} {The centre for
  speech, language and the brain (cslb) concept property norms}.
\newblock \emph{Behavior research methods}, 46.

\bibitem[{Devlin et~al.(2019)Devlin, Chang, Lee, and Toutanova}]{bert}
Jacob Devlin, Ming-Wei Chang, Kenton Lee, and Kristina Toutanova. 2019.
\newblock \href {https://doi.org/10.18653/v1/N19-1423} {{BERT}: Pre-training of
  deep bidirectional transformers for language understanding}.
\newblock In \emph{Proceedings of the 2019 Conference of the North {A}merican
  Chapter of the Association for Computational Linguistics: Human Language
  Technologies, Volume 1 (Long and Short Papers)}, pages 4171--4186,
  Minneapolis, Minnesota. Association for Computational Linguistics.

\bibitem[{Ettinger(2020)}]{BERTisNot}
Allyson Ettinger. 2020.
\newblock \href {https://doi.org/10.1162/tacl_a_00298} {What {BERT} is not:
  Lessons from a new suite of psycholinguistic diagnostics for language
  models}.
\newblock \emph{Transactions of the Association for Computational Linguistics},
  8:34--48.

\bibitem[{Gegenfurtner and Rieger(2000)}]{gegenfurtner-2000-sensory}
Karl~R Gegenfurtner and Jochem Rieger. 2000.
\newblock \href {https://doi.org/https://doi.org/10.1016/S0960-9822(00)00563-7}
  {Sensory and cognitive contributions of color to the recognition of natural
  scenes}.
\newblock \emph{Current Biology}, 10(13):805--808.

\bibitem[{Gladkova et~al.(2016)Gladkova, Drozd, and Matsuoka}]{BATS}
Anna Gladkova, Aleksandr Drozd, and Satoshi Matsuoka. 2016.
\newblock \href {https://doi.org/10.18653/v1/N16-2002} {Analogy-based detection
  of morphological and semantic relations with word embeddings: what works and
  what doesn{'}t.}
\newblock In \emph{Proceedings of the {NAACL} Student Research Workshop}, pages
  8--15, San Diego, California. Association for Computational Linguistics.

\bibitem[{Gordon and Van~Durme(2013)}]{gordon2013reporting}
Jonathan Gordon and Benjamin Van~Durme. 2013.
\newblock Reporting bias and knowledge acquisition.
\newblock In \emph{Proceedings of the 2013 workshop on Automated knowledge base
  construction}, pages 25--30.

\bibitem[{Goyal et~al.(2017)Goyal, Khot, Summers{-}Stay, Batra, and
  Parikh}]{goyal2017vqa}
Yash Goyal, Tejas Khot, Douglas Summers{-}Stay, Dhruv Batra, and Devi Parikh.
  2017.
\newblock \href {https://doi.org/10.1109/CVPR.2017.670} {Making the {V} in
  {VQA} matter: Elevating the role of image understanding in visual question
  answering}.
\newblock In \emph{2017 {IEEE} Conference on Computer Vision and Pattern
  Recognition, {CVPR} 2017, Honolulu, HI, USA, July 21-26, 2017}, pages
  6325--6334. {IEEE} Computer Society.

\bibitem[{Grice(1975)}]{grice1975logic}
Herbert~P Grice. 1975.
\newblock Logic and conversation.
\newblock In \emph{Speech acts}, pages 41--58. Brill.

\bibitem[{Hudson and Manning(2019)}]{hudson2018gqa}
Drew~A. Hudson and Christopher~D. Manning. 2019.
\newblock \href {https://doi.org/10.1109/CVPR.2019.00686} {{GQA:} {A} new
  dataset for real-world visual reasoning and compositional question
  answering}.
\newblock In \emph{{IEEE} Conference on Computer Vision and Pattern
  Recognition, {CVPR} 2019, Long Beach, CA, USA, June 16-20, 2019}, pages
  6700--6709. Computer Vision Foundation / {IEEE}.

\bibitem[{Jia et~al.(2021)Jia, Yang, Xia, Chen, Parekh, Pham, Le, Sung, Li, and
  Duerig}]{align}
Chao Jia, Yinfei Yang, Ye~Xia, Yi-Ting Chen, Zarana Parekh, Hieu Pham, Quoc~V.
  Le, Yunhsuan Sung, Zhen Li, and Tom Duerig. 2021.
\newblock \href {http://arxiv.org/abs/2102.05918} {Scaling up visual and
  vision-language representation learning with noisy text supervision}.

\bibitem[{Jiang et~al.(2020)Jiang, Xu, Araki, and
  Neubig}]{jiang-etal-2020-know}
Zhengbao Jiang, Frank~F. Xu, Jun Araki, and Graham Neubig. 2020.
\newblock \href {https://doi.org/10.1162/tacl_a_00324} {How can we know what
  language models know?}
\newblock \emph{Transactions of the Association for Computational Linguistics},
  8:423--438.

\bibitem[{Kay et~al.(2009)Kay, Berlin, Maffi, Merrifield, and
  Cook}]{kay2009world}
Paul Kay, Brent Berlin, Luisa Maffi, William~R Merrifield, and Richard Cook.
  2009.
\newblock \emph{The world color survey}.
\newblock CSLI Publications Stanford, CA.

\bibitem[{Kingma and Ba(2015)}]{kingma2014adam}
Diederik~P. Kingma and Jimmy Ba. 2015.
\newblock \href {http://arxiv.org/abs/1412.6980} {Adam: {A} method for
  stochastic optimization}.
\newblock In \emph{3rd International Conference on Learning Representations,
  {ICLR} 2015, San Diego, CA, USA, May 7-9, 2015, Conference Track
  Proceedings}.

\bibitem[{Kucera and Francis(1967)}]{browncorpus}
H.~Kucera and W.~N. Francis. 1967.
\newblock \emph{Computational analysis of present-day American English}.
\newblock Brown University Press, Providence, RI.

\bibitem[{Kuznetsova et~al.(2020)Kuznetsova, Rom, Alldrin, Uijlings, Krasin,
  Pont-Tuset, Kamali, Popov, Malloci, Kolesnikov, and et~al.}]{OpenImages}
Alina Kuznetsova, Hassan Rom, Neil Alldrin, Jasper Uijlings, Ivan Krasin, Jordi
  Pont-Tuset, Shahab Kamali, Stefan Popov, Matteo Malloci, Alexander
  Kolesnikov, and et~al. 2020.
\newblock \href {https://doi.org/10.1007/s11263-020-01316-z} {The open images
  dataset v4}.
\newblock \emph{International Journal of Computer Vision}, 128(7):1956–1981.

\bibitem[{Lan et~al.(2020)Lan, Chen, Goodman, Gimpel, Sharma, and
  Soricut}]{albert}
Zhenzhong Lan, Mingda Chen, Sebastian Goodman, Kevin Gimpel, Piyush Sharma, and
  Radu Soricut. 2020.
\newblock \href {https://openreview.net/forum?id=H1eA7AEtvS} {{ALBERT:} {A}
  lite {BERT} for self-supervised learning of language representations}.
\newblock In \emph{8th International Conference on Learning Representations,
  {ICLR} 2020, Addis Ababa, Ethiopia, April 26-30, 2020}. OpenReview.net.

\bibitem[{Lin et~al.(2020)Lin, Lee, Khanna, and Ren}]{NumerSense}
Bill~Yuchen Lin, Seyeon Lee, Rahul Khanna, and Xiang Ren. 2020.
\newblock \href {https://doi.org/10.18653/v1/2020.emnlp-main.557} {{B}irds have
  four legs?! {N}umer{S}ense: {P}robing {N}umerical {C}ommonsense {K}nowledge
  of {P}re-{T}rained {L}anguage {M}odels}.
\newblock In \emph{Proceedings of the 2020 Conference on Empirical Methods in
  Natural Language Processing (EMNLP)}, pages 6862--6868, Online. Association
  for Computational Linguistics.

\bibitem[{Lin et~al.(2019)Lin, Tan, and Frank}]{lin2019open}
Yongjie Lin, Yi~Chern Tan, and Robert Frank. 2019.
\newblock \href {https://doi.org/10.18653/v1/W19-4825} {Open sesame: Getting
  inside {BERT}{'}s linguistic knowledge}.
\newblock In \emph{Proceedings of the 2019 ACL Workshop BlackboxNLP: Analyzing
  and Interpreting Neural Networks for NLP}, pages 241--253, Florence, Italy.
  Association for Computational Linguistics.

\bibitem[{Lin et~al.(2012)Lin, Michel, Aiden~Lieberman, Orwant, Brockman, and
  Petrov}]{googlengrams}
Yuri Lin, Jean-Baptiste Michel, Erez Aiden~Lieberman, Jon Orwant, Will
  Brockman, and Slav Petrov. 2012.
\newblock \href {https://aclanthology.org/P12-3029} {Syntactic annotations for
  the {G}oogle {B}ooks {NG}ram corpus}.
\newblock In \emph{Proceedings of the {ACL} 2012 System Demonstrations}, pages
  169--174, Jeju Island, Korea. Association for Computational Linguistics.

\bibitem[{Liu et~al.(2019)Liu, Ott, Goyal, Du, Joshi, Chen, Levy, Lewis,
  Zettlemoyer, and Stoyanov}]{roberta}
Yinhan Liu, Myle Ott, Naman Goyal, Jingfei Du, Mandar Joshi, Danqi Chen, Omer
  Levy, Mike Lewis, Luke Zettlemoyer, and Veselin Stoyanov. 2019.
\newblock \href {https://arxiv.org/abs/1907.11692} {Roberta: A robustly
  optimized bert pretraining approach}.
\newblock \emph{ArXiv preprint}, abs/1907.11692.

\bibitem[{Misra et~al.(2016)Misra, Zitnick, Mitchell, and
  Girshick}]{mirsha2016seeing}
Ishan Misra, C.~Lawrence Zitnick, Margaret Mitchell, and Ross~B. Girshick.
  2016.
\newblock \href {https://doi.org/10.1109/CVPR.2016.320} {Seeing through the
  human reporting bias: Visual classifiers from noisy human-centric labels}.
\newblock In \emph{2016 {IEEE} Conference on Computer Vision and Pattern
  Recognition, {CVPR} 2016, Las Vegas, NV, USA, June 27-30, 2016}, pages
  2930--2939. {IEEE} Computer Society.

\bibitem[{Nelson et~al.(2004)Nelson, McEvoy, and
  Schreiber}]{nelson-2004-university}
Douglas~L Nelson, Cathy~L McEvoy, and Thomas~A Schreiber. 2004.
\newblock The university of south florida free association, rhyme, and word
  fragment norms.
\newblock \emph{Behavior Research Methods, Instruments, \& Computers},
  36(3):402--407.

\bibitem[{Petroni et~al.(2019)Petroni, Rockt{\"a}schel, Riedel, Lewis, Bakhtin,
  Wu, and Miller}]{petroni2019language}
Fabio Petroni, Tim Rockt{\"a}schel, Sebastian Riedel, Patrick Lewis, Anton
  Bakhtin, Yuxiang Wu, and Alexander Miller. 2019.
\newblock \href {https://doi.org/10.18653/v1/D19-1250} {Language models as
  knowledge bases?}
\newblock In \emph{Proceedings of the 2019 Conference on Empirical Methods in
  Natural Language Processing and the 9th International Joint Conference on
  Natural Language Processing (EMNLP-IJCNLP)}, pages 2463--2473, Hong Kong,
  China. Association for Computational Linguistics.

\bibitem[{Radford et~al.(2021)Radford, Kim, Hallacy, Ramesh, Goh, Agarwal,
  Sastry, Askell, Mishkin, Clark, Krueger, and Sutskever}]{clip}
Alec Radford, Jong~Wook Kim, Chris Hallacy, Aditya Ramesh, Gabriel Goh,
  Sandhini Agarwal, Girish Sastry, Amanda Askell, Pamela Mishkin, Jack Clark,
  Gretchen Krueger, and Ilya Sutskever. 2021.
\newblock \href {http://arxiv.org/abs/2103.00020} {Learning transferable visual
  models from natural language supervision}.

\bibitem[{Radford et~al.(2019)Radford, Wu, Child, Luan, Amodei, and
  Sutskever}]{gpt2}
Alec Radford, Jeffrey Wu, Rewon Child, David Luan, Dario Amodei, and Ilya
  Sutskever. 2019.
\newblock Language models are unsupervised multitask learners.
\newblock \emph{OpenAI blog}, 1(8):9.

\bibitem[{Rosenthal et~al.(2018)Rosenthal, Ratnasingam, Haile, Eastman,
  Fuller-Deets, and Conway}]{rosenthal-2018-color}
Isabelle Rosenthal, Sivalogeswaran Ratnasingam, Theodros Haile, Serena Eastman,
  Josh Fuller-Deets, and Bevil~R. Conway. 2018.
\newblock \href {https://doi.org/10.1167/18.11.1} {{Color statistics of
  objects, and color tuning of object cortex in macaque monkey}}.
\newblock \emph{Journal of Vision}, 18(11):1--1.

\bibitem[{Shwartz and Choi(2020)}]{shwartz2020neural}
Vered Shwartz and Yejin Choi. 2020.
\newblock \href {https://doi.org/10.18653/v1/2020.coling-main.605} {Do neural
  language models overcome reporting bias?}
\newblock In \emph{Proceedings of the 28th International Conference on
  Computational Linguistics}, pages 6863--6870, Barcelona, Spain (Online).
  International Committee on Computational Linguistics.

\bibitem[{Tan and Bansal(2020)}]{vokenization}
Hao Tan and Mohit Bansal. 2020.
\newblock \href {https://doi.org/10.18653/v1/2020.emnlp-main.162}
  {Vokenization: Improving language understanding with contextualized,
  visual-grounded supervision}.
\newblock In \emph{Proceedings of the 2020 Conference on Empirical Methods in
  Natural Language Processing (EMNLP)}, pages 2066--2080, Online. Association
  for Computational Linguistics.

\bibitem[{Teichmann et~al.(2020)Teichmann, Quek, Robinson, Grootswagers,
  Carlson, and Rich}]{teichmann2020influence}
Lina Teichmann, Genevieve~L. Quek, Amanda~K. Robinson, Tijl Grootswagers,
  Thomas~A. Carlson, and Anina~N. Rich. 2020.
\newblock \href {https://doi.org/10.1523/JNEUROSCI.0158-20.2020} {The influence
  of object-color knowledge on emerging object representations in the brain}.
\newblock \emph{Journal of Neuroscience}, 40(35):6779--6789.

\bibitem[{Torralba and Efros(2011)}]{torralba2011}
Antonio Torralba and Alexei~A. Efros. 2011.
\newblock \href {https://doi.org/10.1109/CVPR.2011.5995347} {Unbiased look at
  dataset bias}.
\newblock In \emph{The 24th {IEEE} Conference on Computer Vision and Pattern
  Recognition, {CVPR} 2011, Colorado Springs, CO, USA, 20-25 June 2011}, pages
  1521--1528. {IEEE} Computer Society.

\bibitem[{van Miltenburg(2016)}]{emiel2018stereotyping}
Emiel van Miltenburg. 2016.
\newblock \href {https://arxiv.org/abs/1605.06083} {Stereotyping and bias in
  the flickr30k dataset}.
\newblock \emph{ArXiv preprint}, abs/1605.06083.

\bibitem[{Voita and Titov(2020)}]{voitamdl}
Elena Voita and Ivan Titov. 2020.
\newblock \href {https://doi.org/10.18653/v1/2020.emnlp-main.14}
  {Information-theoretic probing with minimum description length}.
\newblock In \emph{Proceedings of the 2020 Conference on Empirical Methods in
  Natural Language Processing (EMNLP)}, pages 183--196, Online. Association for
  Computational Linguistics.

\bibitem[{Warstadt et~al.(2020)Warstadt, Parrish, Liu, Mohananey, Peng, Wang,
  and Bowman}]{blimp}
Alex Warstadt, Alicia Parrish, Haokun Liu, Anhad Mohananey, Wei Peng, Sheng-Fu
  Wang, and Samuel~R. Bowman. 2020.
\newblock \href {https://doi.org/10.1162/tacl_a_00321} {{BL}i{MP}: The
  benchmark of linguistic minimal pairs for {E}nglish}.
\newblock \emph{Transactions of the Association for Computational Linguistics},
  8:377--392.

\bibitem[{Weir et~al.(2020)Weir, Poliak, and Van~Durme}]{weir2020probing}
Nathaniel Weir, Adam Poliak, and Benjamin Van~Durme. 2020.
\newblock Probing neural language models for human tacit assumptions.
\newblock CogSci.

\bibitem[{Whitney et~al.(2021)Whitney, Song, Brandfonbrener, Altosaar, and
  Cho}]{whitney2021evaluating}
William~F. Whitney, Min~Jae Song, David Brandfonbrener, Jaan Altosaar, and
  Kyunghyun Cho. 2021.
\newblock \href {http://arxiv.org/abs/2009.07368} {Evaluating representations
  by the complexity of learning low-loss predictors}.

\bibitem[{Wolf et~al.(2019)Wolf, Debut, Sanh, Chaumond, Delangue, Moi, Cistac,
  Rault, Louf, Funtowicz, Davison, Shleifer, von Platen, Ma, Jernite, Plu, Xu,
  Scao, Gugger, Drame, Lhoest, and Rush}]{huggingface}
Thomas Wolf, Lysandre Debut, Victor Sanh, Julien Chaumond, Clement Delangue,
  Anthony Moi, Pierric Cistac, Tim Rault, Rémi Louf, Morgan Funtowicz, Joe
  Davison, Sam Shleifer, Patrick von Platen, Clara Ma, Yacine Jernite, Julien
  Plu, Canwen Xu, Teven~Le Scao, Sylvain Gugger, Mariama Drame, Quentin Lhoest,
  and Alexander~M. Rush. 2019.
\newblock \href {https://arxiv.org/abs/1910.03771} {Huggingface's transformers:
  State-of-the-art natural language processing}.
\newblock \emph{ArXiv preprint}, abs/1910.03771.

\bibitem[{Zhang and Bowman(2018)}]{zhang-bowman-2018-language}
Kelly Zhang and Samuel Bowman. 2018.
\newblock \href {https://doi.org/10.18653/v1/W18-5448} {Language modeling
  teaches you more than translation does: Lessons learned through auxiliary
  syntactic task analysis}.
\newblock In \emph{Proceedings of the 2018 {EMNLP} Workshop {B}lackbox{NLP}:
  Analyzing and Interpreting Neural Networks for {NLP}}, pages 359--361,
  Brussels, Belgium. Association for Computational Linguistics.

\end{thebibliography}
\bibliographystyle{acl_natbib}

\appendix

\section{Dataset Construction}
\subsection{Analysis of Annotator Biases}
\label{app:annotator_biases}
A potential side-effect of crowdsourcing annotations is that annotators might be biased toward choosing fewer colors faster, as this would equate to higher monetary incentives. 
We observe a small correlation (Kendall’s Tau=0.154, p=0.026) between the total time and number of colors selected. 
However, this is to be expected as selecting the colors takes time. 

All models we evaluate were predominately trained on English text. To accommodate this domain and minimize dataset variance, we recruit only annotators from the United States. This may induce cultural or geographic biases: e.g., the color diversity of carrots is much smaller in the United States than in some Asian countries. Other geographic biases are more fine-grained; for example, the color of fire hydrants in the U.S. depends on where you live and the water source.

Additionally, our choice of colors is not as universal as, for example, the 6 color terms defined by The World Color Survey \cite{kay2009world}. The latter may be more suitable for multilingual studies, though we leave such investigations for future work.

\section{Experimental Details}
For all experiments, we implement the CoDa dataset using the Huggingface Datasets Library. We use Huggingface's \citep{huggingface} pretrained models for evaluating all text-only models, and the official CLIP implementation by \citet{clip} for all CLIP models.\footnote{\href{https://github.com/openai/CLIP}{github.com/openai/CLIP}} We run all experiments on a single machine with one Nvidia Titan RTX GPU. 

\subsection{Representation Probing}
\label{app:reprprobinghyperparams}
Our representation probing implementation is derived from the efficient JAX version provided by \citet{whitney2021evaluating}.\footnote{\href{https://github.com/willwhitney/reprieve}{github.com/willwhitney/reprieve}} 

We split the training set into 10 subsets spaced logarithmically from 1 to 311 objects, and report averages over 5 seeds. Note that for each seed, any additional points along the curve represent additional objects to the previous subset, however, different seeds have different object sets and thus a different number of samples per subset. For our dataset, we found the difference in samples to be far less impactful on performance than the number of objects.

All probes are 2-layer MLPs with ReLU activation functions and are trained using the Adam Optimizer \cite{kingma2014adam} with a learning rate of $10^{-4}$. All probes are trained for 4000 steps. More details on how to reproduce the experiments are provided in our GitHub repository.\footnote{\href{https://github.com/nala-cub/coda}{github.com/nala-cub/coda}}

\section{Zero Shot Results}
\label{app:zeroshotresults}
The zero-shot results for all evaluated LMs are provided in \cref{tab:lmzeroshot:all}. 
\begin{table*}[th!]\centering\small
\begin{tabular}{rllcccrrr}
\toprule
\mcb{2}{Model} & \mcb{1}{Group} & \mcb{1}{Spearman $\rho \uparrow$} &   \mcb{1}{Kendall's $\tau \uparrow$} &  \mcb{1}{Acc@1 $\uparrow$} &  \mcb{1}{$\mathbf{D_{JS}} \downarrow$} &
$\Delta \rho \uparrow$ & $\Delta \tau \uparrow$\\
\midrule
\mrvert{12}{GPT-2} & B & Single &      \flatpm{34.9}{25.7} &       \flatpm{28.8}{21.2} &           $26.8$ &  \flatpm{0.39}{0.07} &                $-6.86$ &                $-6.60$ \\
        &     & Multi &      \flatpm{40.7}{22.9} &       \flatpm{32.9}{18.5} &           $23.6$ &  \flatpm{0.25}{0.06} &                $-6.77$ &                $-5.46$ \\
        &     & Any &      \boldpm{45.7}{25.0} &       \boldpm{36.9}{20.9} &  $\textbf{33.9}$ &  \boldpm{0.09}{0.04} &        $\textbf{2.17}$ &        $\textbf{2.64}$ \\
        & M & Single &      \flatpm{34.2}{26.4} &       \flatpm{28.5}{21.8} &  $\textbf{35.9}$ &  \flatpm{0.39}{0.07} &                $-6.94$ &                $-6.30$ \\
        &     & Multi &      \flatpm{36.0}{26.0} &       \flatpm{29.5}{20.8} &           $25.5$ &  \flatpm{0.25}{0.06} &               $-10.87$ &                $-8.51$ \\
        &     & Any &      \boldpm{43.2}{26.5} &       \boldpm{34.8}{21.2} &           $35.7$ &  \boldpm{0.09}{0.04} &        $\textbf{0.09}$ &        $\textbf{0.74}$ \\
        & L & Single &      \flatpm{39.9}{24.4} &       \flatpm{32.8}{19.7} &           $33.8$ &  \flatpm{0.39}{0.07} &                $-1.10$ &                $-1.91$ \\
        &     & Multi &      \flatpm{44.8}{20.9} &       \flatpm{36.5}{16.8} &           $29.8$ &  \flatpm{0.26}{0.06} &                $-1.49$ &                $-1.05$ \\
        &     & Any &      \boldpm{47.3}{26.9} &       \boldpm{37.9}{21.8} &  $\textbf{38.3}$ &  \boldpm{0.09}{0.04} &        $\textbf{4.35}$ &        $\textbf{3.95}$ \\
        & XL & Single &      \flatpm{40.3}{26.6} &       \flatpm{33.6}{22.1} &  $\textbf{40.4}$ &  \flatpm{0.39}{0.07} &                $-0.55$ &                $-1.01$ \\
        &     & Multi &      \flatpm{41.7}{24.3} &       \flatpm{34.1}{19.4} &           $28.8$ &  \flatpm{0.25}{0.06} &                $-4.66$ &                $-3.42$ \\
        &     & Any &      \boldpm{48.1}{25.1} &       \boldpm{38.2}{20.2} &           $40.0$ &  \boldpm{0.09}{0.04} &        $\textbf{5.29}$ &        $\textbf{4.46}$ \\
\midrule
\mrvert{6}{RoBERTa} & B & Single &      \flatpm{41.5}{23.9} &       \flatpm{34.4}{19.6} &  $\textbf{32.3}$ &  \flatpm{0.32}{0.13} &                 $0.58$ &                $-0.21$ \\
        &     & Multi &      \flatpm{47.0}{21.9} &       \flatpm{37.7}{18.0} &           $23.1$ &  \flatpm{0.21}{0.09} &                 $0.44$ &                $-0.03$ \\
        &     & Any &      \boldpm{51.9}{22.7} &       \boldpm{41.3}{18.9} &           $29.6$ &  \boldpm{0.11}{0.07} &        $\textbf{8.64}$ &        $\textbf{7.27}$ \\
        & L & Single &      \flatpm{47.8}{24.7} &       \flatpm{40.1}{20.8} &  $\textbf{42.9}$ &  \flatpm{0.28}{0.11} &                 $7.17$ &                 $5.69$ \\
        &     & Multi &      \flatpm{50.2}{23.8} &       \flatpm{41.0}{19.5} &           $33.2$ &  \flatpm{0.19}{0.08} &                 $4.57$ &                 $4.01$ \\
        &     & Any &      \boldpm{52.5}{23.5} &       \boldpm{42.0}{19.5} &           $36.5$ &  \boldpm{0.10}{0.06} &        $\textbf{9.97}$ &        $\textbf{8.26}$ \\
\midrule
\mrvert{12}{ALBERT V1} & B & Single &      \flatpm{27.8}{25.0} &       \flatpm{23.2}{20.3} &           $16.2$ &  \flatpm{0.38}{0.10} &               $-14.08$ &               $-12.30$ \\
        &     & Multi &      \flatpm{31.4}{24.2} &       \flatpm{25.1}{18.8} &           $13.0$ &  \flatpm{0.27}{0.09} &               $-15.27$ &               $-12.60$ \\
        &     & Any &      \boldpm{42.8}{26.4} &       \boldpm{33.7}{21.4} &  $\textbf{18.3}$ &  \boldpm{0.14}{0.06} &       $\textbf{-0.70}$ &       $\textbf{-0.63}$ \\
        & L & Single &      \flatpm{29.4}{27.0} &       \flatpm{24.3}{21.9} &           $31.8$ &  \flatpm{0.35}{0.13} &               $-11.67$ &               $-10.62$ \\
        &     & Multi &      \flatpm{32.7}{22.6} &       \flatpm{26.7}{18.0} &           $23.1$ &  \flatpm{0.25}{0.09} &               $-13.50$ &               $-10.78$ \\
        &     & Any &      \boldpm{41.2}{25.7} &       \boldpm{33.6}{20.5} &  $\textbf{38.3}$ &  \boldpm{0.13}{0.06} &       $\textbf{-2.37}$ &       $\textbf{-0.58}$ \\
        & XL & Single &      \flatpm{36.4}{24.5} &       \flatpm{29.7}{20.0} &           $26.3$ &  \flatpm{0.35}{0.11} &                $-4.73$ &                $-4.99$ \\
        &     & Multi &      \flatpm{44.6}{19.1} &       \flatpm{36.1}{15.5} &           $26.9$ &  \flatpm{0.22}{0.07} &                $-1.53$ &                $-1.27$ \\
        &     & Any &      \boldpm{48.2}{21.4} &       \boldpm{38.2}{17.2} &  $\textbf{35.7}$ &  \boldpm{0.11}{0.05} &        $\textbf{5.07}$ &        $\textbf{4.22}$ \\
        & XXL & Single &      \flatpm{39.9}{25.6} &       \flatpm{33.1}{21.1} &           $31.3$ &  \flatpm{0.31}{0.12} &                $-1.38$ &                $-1.80$ \\
        &     & Multi &      \flatpm{41.3}{26.1} &       \boldpm{33.2}{21.0} &           $23.6$ &  \flatpm{0.21}{0.08} &                $-5.23$ &                $-4.48$ \\
        &     & Any &      \boldpm{41.9}{24.3} &       \flatpm{32.8}{18.4} &  $\textbf{38.3}$ &  \boldpm{0.11}{0.05} &       $\textbf{-0.87}$ &       $\textbf{-1.03}$ \\
\midrule
\mrvert{12}{ALBERT V2} & B & Single &      \flatpm{22.3}{29.7} &       \flatpm{18.9}{24.2} &           $20.7$ &  \flatpm{0.36}{0.11} &               $-19.54$ &               $-16.46$ \\
        &     & Multi &      \flatpm{22.2}{26.8} &       \flatpm{18.0}{21.3} &           $18.3$ &  \flatpm{0.26}{0.07} &               $-23.57$ &               $-19.11$ \\
        &     & Any &      \boldpm{25.8}{26.9} &       \boldpm{20.8}{20.6} &  $\textbf{26.1}$ &  \boldpm{0.12}{0.05} &      $\textbf{-18.38}$ &      $\textbf{-13.98}$ \\
        & L & Single &      \flatpm{39.2}{27.1} &       \flatpm{32.5}{22.4} &           $30.3$ &  \flatpm{0.32}{0.11} &       $\textbf{-2.14}$ &                $-2.50$ \\
        &     & Multi &      \boldpm{41.9}{24.9} &       \boldpm{33.9}{20.4} &           $25.0$ &  \flatpm{0.21}{0.07} &                $-3.73$ &                $-3.10$ \\
        &     & Any &      \flatpm{40.0}{22.9} &       \flatpm{32.4}{18.1} &  $\textbf{33.0}$ &  \boldpm{0.10}{0.05} &                $-3.70$ &       $\textbf{-2.05}$ \\
        & XL & Single &      \flatpm{25.2}{26.6} &       \flatpm{20.5}{21.5} &  $\textbf{26.3}$ &  \flatpm{0.35}{0.12} &               $-16.14$ &               $-14.53$ \\
        &     & Multi &      \flatpm{25.4}{23.4} &       \flatpm{20.7}{18.4} &           $23.6$ &  \flatpm{0.25}{0.08} &               $-20.51$ &               $-16.59$ \\
        &     & Any &      \boldpm{29.6}{27.1} &       \boldpm{23.1}{20.8} &           $26.1$ &  \boldpm{0.12}{0.05} &      $\textbf{-12.48}$ &      $\textbf{-10.04}$ \\
        & XXL & Single &      \flatpm{43.7}{24.4} &       \boldpm{36.4}{20.6} &           $34.3$ &  \flatpm{0.30}{0.11} &        $\textbf{2.69}$ &        $\textbf{1.55}$ \\
        &     & Multi &      \boldpm{45.2}{23.7} &       \flatpm{36.1}{19.5} &           $25.5$ &  \flatpm{0.20}{0.07} &                $-1.08$ &                $-1.40$ \\
        &     & Any &      \flatpm{43.7}{24.9} &       \flatpm{34.5}{19.6} &  $\textbf{39.1}$ &  \boldpm{0.10}{0.05} &                 $0.95$ &                 $0.70$ \\
\midrule
\mcb{2}{Average}  & Single &      \flatpm{35.9}{25.8} &       \flatpm{29.8}{21.2} &           $30.7$ &  \flatpm{0.35}{0.10} &                $-5.33$ &                $-5.14$ \\
        &     & Multi &      \flatpm{38.9}{23.6} &       \flatpm{31.5}{19.0} &           $24.5$ &  \flatpm{0.24}{0.07} &                $-7.37$ &                $-5.99$ \\
        &     & Any &      \boldpm{43.0}{25.0} &       \boldpm{34.3}{19.9} &  $\textbf{33.5}$ &  \boldpm{0.11}{0.05} &       $\textbf{-0.14}$ &        $\textbf{0.28}$ \\
\bottomrule
\end{tabular}

\caption{LM results when probed in a zero-shot setting. Single, Multi, and Any indicate sets of objects that are frequently a singe color, between two to four colors, or could be any color, respectively. All correlation coefficients ($\rho, \tau$) are multiplied by 100. Means and standard deviations are calculated over objects of the respective group.  
}
\label{tab:lmzeroshot:all}
\end{table*}

\end{document}